\documentclass[letterpaper, 10 pt, conference]{ieeeconf}

\IEEEoverridecommandlockouts
\makeatletter
\let\NAT@parse\undefined
\makeatother

    \usepackage{cite}
    \usepackage{amsmath}
    \usepackage{amsthm}
    \usepackage{amsfonts}
    \usepackage{amssymb}
    \usepackage{graphicx}
    \usepackage{array}
    \usepackage[dvipsnames]{xcolor}
    \usepackage{subcaption}
    \usepackage{mathrsfs}
    \usepackage{multicol}
    \usepackage{multirow}
    \usepackage{arydshln}
    \usepackage{diagbox}
    
\usepackage{duckuments}
\usepackage{amsmath}
\usepackage{tabularx}
\usepackage{booktabs}
\usepackage{makecell}
\usepackage{array}

\usepackage{colortbl}
\usepackage{tikz}
\usepackage{array}
\usepackage{graphicx}
\usepackage{multicol}

\usepackage[bookmarks=true]{hyperref}

        \newcommand{\set}[1]{\left\{#1\right\}}

        \newcommand{\mc}[1]{\mathcal{#1}}

    \usepackage{letltxmacro}
    \LetLtxMacro\orgvdots\vdots
    \LetLtxMacro\orgddots\ddots

    \makeatletter
    \DeclareRobustCommand\vdots{%
        \mathpalette\@vdots{}%
    }
    \newcommand*{\@vdots}[2]{%
        \sbox0{$#1\cdotp\cdotp\cdotp\m@th$}%
        \sbox2{$#1.\m@th$}%
        \vbox{%
            \dimen@=\wd0 %
            \advance\dimen@ -3\ht2 %
            \kern.5\dimen@
            \dimen@=\wd2 %
            \advance\dimen@ -\ht2 %
            \dimen2=\wd0 %
            \advance\dimen2 -\dimen@
            \vbox to \dimen2{%
                \offinterlineskip
                \copy2 \vfill\copy2 \vfill\copy2 %
            }%
        }%
    }
    \DeclareRobustCommand\ddots{%
        \mathinner{%
            \mathpalette\@ddots{}%
            \mkern\thinmuskip
        }%
    }
    \newcommand*{\@ddots}[2]{%
        \sbox0{$#1\cdotp\cdotp\cdotp\m@th$}%
        \sbox2{$#1.\m@th$}%
        \vbox{%
            \dimen@=\wd0 %
            \advance\dimen@ -3\ht2 %
            \kern.5\dimen@
            \dimen@=\wd2 %
            \advance\dimen@ -\ht2 %
            \dimen2=\wd0 %
            \advance\dimen2 -\dimen@
            \vbox to \dimen2{%
                \offinterlineskip
                \hbox{$#1\mathpunct{.}\m@th$}%
                \vfill
                \hbox{$#1\mathpunct{\kern\wd2}\mathpunct{.}\m@th$}%
                \vfill
                \hbox{$#1\mathpunct{\kern\wd2}\mathpunct{\kern\wd2}\mathpunct{.}\m@th$}%
            }%
        }%
    }
    \makeatother

\usepackage{cleveref}
\Crefname{figure}{Fig.}{Figs.}
\Crefname{equation}{Eq.}{Eqs.}
\Crefname{lemma}{Lemma}{Lemmata}
\Crefname{proposition}{Proposition}{Propositions}
\Crefname{assumption}{Assumption}{Assumptions}
\Crefname{theorem}{Theorem}{Theorems}
\Crefname{section}{Section}{Sections}
\Crefname{subsection}{Subsection}{Subsections}
\Crefname{appendix}{Appendix}{Appendices}
\Crefname{corollary}{Corollary}{Corollaries}

\DeclareMathOperator{\SO}{SO}
\DeclareMathOperator{\SE}{SE}

\newcommand{\monogram}[3]{{}^{#2}\!#1^{#3}}

\DeclareMathOperator{\Unif}{Unif}

\newcommand\blfootnote[1]{%
  \begingroup
  \renewcommand\thefootnote{}%
  \footnotetext{#1}%
  \addtocounter{footnote}{-1}%
  \endgroup
}

\newcolumntype{Y}{>{\centering\arraybackslash}X}

\usepackage{tikz}
\usetikzlibrary{calc}

\definecolor{mygreen}{RGB}{144,238,144}
\definecolor{myyellow}{RGB}{255,255,110}
\definecolor{myorange}{RGB}{255,179,102}
\definecolor{myred}{RGB}{255,102,102}

\newcommand{\successbarci}[6][4.2]{%
  \begingroup
  \def\W{#1} %
  \def\a{#2}\def\b{#3}\def\c{#4}\def\d{#5}%
  \def\ci{#6} %
  \pgfmathsetmacro{\total}{\a+\b+\c+\d}%
  \pgfmathsetmacro{\pa}{\a/\total}%
  \pgfmathsetmacro{\pb}{\b/\total}%
  \pgfmathsetmacro{\pc}{\c/\total}%
  \pgfmathsetmacro{\pd}{\d/\total}%
  \def\H{0.4} %

  \begin{tikzpicture}[baseline=(current bounding box.center)]
    \coordinate (L) at (0,0);

    \pgfmathsetmacro{\w}{\pa*\W}
    \coordinate (R) at ($(L)+(\w,0)$);
    \fill[mygreen,draw=black] (L) rectangle ($(R)+(0,\H)$);
    \ifx\ci\empty
      \node[below=-2pt] at ($ (L)!0.5!(R) $) {\scriptsize \a};
    \else
      \node[below=-2pt] at ($ (L)!0.5!(R) $) {\scriptsize $\a \pm \ci$};
    \fi
    \coordinate (L) at (R);

    \foreach \p/\val/\col in {\pb/\b/myyellow,\pc/\c/myorange,\pd/\d/myred} {
        \pgfmathsetmacro{\w}{\p*\W}
        \coordinate (R) at ($(L)+(\w,0)$);
        \fill[\col,draw=black] (L) rectangle ($(R)+(0,\H)$);
        \node[below=-2pt] at ($ (L)!0.5!(R) $) {\scriptsize \val};
        \coordinate (L) at (R);
    }
  \end{tikzpicture}%
  \endgroup
}

\usepackage{color-edits}
\usepackage{enumerate}%

\definecolor{darkgreen}{rgb}{0.0, 0.75, 0.0}
\addauthor{lf}{blue}
\addauthor{aw}{orange}
\addauthor{tc}{cyan}
\addauthor{np}{darkgreen}
\addauthor{bc}{purple}
\addauthor{rt}{red}

\title{\LARGE \bf
How Well do Diffusion Policies Learn Kinematic Constraint Manifolds?
}

\author{Lexi Foland, Thomas Cohn, Adam Wei, Nicholas Pfaff, Boyuan Chen, and Russ Tedrake}

\begin{document}

\twocolumn[{
\renewcommand\twocolumn[1][]{#1}
\maketitle
\includegraphics[width=\linewidth, keepaspectratio]{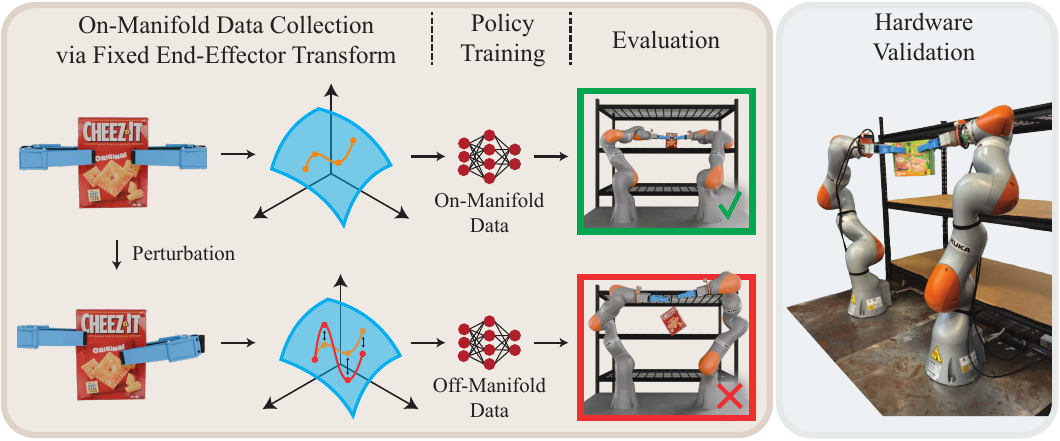}
\vspace{-10pt}
{
\captionsetup{hypcap=false}
\captionof{figure}{
\textbf{An overview of our method.} We collect teleoperation data for a constrained bimanual pick-and-place task. Then, we perturb these demonstrations to generate three additional datasets that still accomplish the task, but contain increasing constraint violation. We train a policy on each of these datasets and analyze task success and constraint adherence. Lastly, we collect demonstrations for the same task on hardware, train a policy, and evaluate its performance on similar metrics.
\label{fig:teaser}
}
}
\vspace{1em}

}]

\thispagestyle{empty}
\pagestyle{empty}

\begin{abstract}
Diffusion policies have shown impressive results in robot imitation learning, even for tasks that require satisfaction of kinematic equality constraints.
However, task performance alone is not a reliable indicator of the policy's ability to precisely learn constraints in the training data.
To investigate, we analyze how well diffusion policies discover these manifolds with a case study on a bimanual pick-and-place task that encourages fulfillment of a kinematic constraint for success.
We study how three factors affect trained policies: dataset size, dataset quality, and manifold curvature.
Our experiments show diffusion policies learn a coarse approximation of the constraint manifold with learning affected negatively by decreases in both dataset size and quality. On the other hand, the curvature of the constraint manifold showed inconclusive correlations with both constraint satisfaction and task success. A hardware evaluation verifies the applicability of our results in the real world.
Project website with additional results and visuals: \url{https://diffusion-learns-kinematic.github.io/}.

\end{abstract}

\blfootnote{This work was supported by Amazon.com, PO No. 2D-15694048, the Office of Naval Research, PO No. N000142412603, SRI International, PO No. PO81455, the Defense Science and Technology Agency, PO No. DST00OECI20300823, the NSERC CGS D-587703, and the National Science Foundation Graduate Research Fellowship Program under Grant No. 2141064. Any opinions, findings, and conclusions or recommendations expressed in this material are those of the author(s) and do not necessarily reflect the views of the National Science Foundation (or other funding organizations). The authors are with the Computer Science and Artificial Intelligence Laboratory (CSAIL), Massachusetts Institute of Technology, Cambridge, Massachusetts. Corresponding author: \texttt{lkfoland@mit.edu}}%

\section{Introduction}
\label{sec:intro}

Advances in generative modeling have laid the foundation for several breakthroughs in robot imitation learning. For example, diffusion and flow-based policies have shown that dexterous and even kinematically constrained manipulation tasks can be learned from just a few hours of robot data \cite{chi2023diffusion,black2025pi05,trilbmteam2025carefulexaminationlargebehavior}.
A central idea in generative modeling is the \textit{manifold hypothesis}, which posits that high-dimensional real-world datasets are concentrated along lower-dimensional manifolds in the higher-dimensional space~\cite{bengio2013representationlearning,fefferman2013testingmanifoldhypothesis}.
If one accepts this hypothesis, then diffusion is approximately equivalent to Euclidean projections onto the data manifold \cite{permenter2024interpretingimprovingdiffusionmodels}. By extension, Diffusion Policy \cite{chi2023diffusion} can be interpreted as a method for sampling from the manifold of robot trajectories.

The data manifold in robotics is challenging to characterize; however, robot trajectories often exhibit well-understood structures. For instance, robot motion frequently operates along a submanifold of configuration space defined by kinematic constraints. While these constraints are implicitly present in datasets, they are rarely encoded explicitly during training. Thus, there is no guarantee that the trained policies will respect these constraints, even when they may be critical for task completion. Furthermore, policies are often deployed in conjunction with constraint-enforcing or compliant low-level controllers that mitigate constraint violations in the policy’s action predictions. Thus, there is value in quantifying the extent to which diffusion policies adhere to the constraint manifold in robot data (independent of the lower-level control stack), as well as carefully studying the factors that affect the quality of this adherence.

In this paper, we study how well diffusion policies learn the kinematic constraint manifolds in robot data. We consider a bimanual pick-and-place problem where a robot must move objects from a table into a shelf while avoiding collisions with the environment. Maintaining a constant relative pose between the grippers is essential to avoid dropping the object during transport. This requirement is a nonlinear kinematic constraint in the robot’s configuration space. We collect data for this task using a specialized teleoperation setup that allows the teleoperator to lock the relative transform between the end-effectors during object transport. This ensures all training data satisfies the constraint and also provides a more intuitive and forgiving user experience. \Cref{fig:teaser} contains images of the simulation and hardware setups as well as a broad overview of our method.

While there are a variety of factors that affect policy performance, we isolate and study the policy's adherence to kinematic constraints, independent of confounding factors such as choice of low-level controller, gripper compliance, and contact forces.
More concretely, we study the effect of
\begin{enumerate}
    \item \textbf{Data amount:} by training policies with varying numbers of demonstrations.
    \item \textbf{Data quality:} by artificially introducing constraint violations into the training data. Although the demonstrations still succeed at the task, the more the policy learns that it can violate the constraints, the more likely it is to drop the box.
    \item \textbf{Constraint difficulty:} by measuring the curvature of the kinematic constraint manifold.
\end{enumerate}

We coarsely evaluate policies on an outcome basis (success, partial success, partial failure, and full failure).
Importantly, we also measure the policy's ability to learn the constraints by evaluating the action predictions.
Considering the predictions instead of the measured positions isolates the policy's constraint satisfaction from confounding factors, such as the low-level controller and contact forces. Lastly, we consider correlations between constraint manifold curvature, task success, and constraint satisfaction. We conclude by conducting experiments on hardware to verify that our methods and experiments translate to the real world.

\section{Related Work}
\label{sec:related_work}
One interpretation of the diffusion process, and generative modeling more broadly, is sampling from the data manifold \cite{ho2020denoisingdiffusionprobabilisticmodels, permenter2024interpretingimprovingdiffusionmodels}. The problem of sampling from a subset of this manifold (described by explicit constraints or a reward function) has been studied in both the generative modelling and robotics literature. Techniques generally fall under two categories: \emph{explicit} methods and \emph{guidance}-based methods.

Explicit methods allow diffusion models to sample exactly from some predefined manifold.
But these methods are often specialized to certain well-understood manifolds, such as $\SE(3)$~\cite{urain2023se} or the unit sphere~\cite{li2025adpro}.
More general methods still require efficient implementations of complex mathematical operations like the exponential map~\cite{debortoli2022riemannianscorebasedgenerativemodelling,fishman2023metropolis}, geodesic distance~\cite{chen2024flow}, or parallel transport~\cite{fishman2024diffusionmodelsconstraineddomains}.
Such computations must be numerically approximated for many of the constraint manifolds of interest in robotics, rendering these approaches intractable.
Alternative methods, such as motion primitive diffusion~\cite{scheikl2024movement}, use an existing movement primitive network within the diffusion framework to ensure constraint satisfaction; but this requires the user to pre-specify primitives and precludes the simple end-to-end training pipeline that has made Diffusion Policy so popular.

Alternatively, guidance-based methods, which use an extra term to bias sampling, can approximately sample from a desired subset of the manifold or distribution. For instance, classifier guidance uses gradients from an external classifier \cite{dhariwal2021diffusionmodelsbeatgans}, while classifier-free guidance simulates this gradient by interpolating between conditional and unconditional scores \cite{ho2022classifierfreediffusionguidance}. In robotics and offline RL, gradients from a reward function \cite{janner2022planningdiffusionflexiblebehavior, ajay2023conditionalgenerativemodelingneed, romer2025diffusion, li2024constraint} or score estimates from a separate diffusion model \cite{wang2024pocopolicycompositionheterogeneous, power2023sampling} can steer sampling towards high reward. While these approaches have seen success, they face limitations: they require the constraint set or reward to be expressable and differentiable, the strength of the guidance needs to be tuned, and they provide no guarantees that the final samples will satisfy the constraints or have high reward.

More commonly in robotics, the constraint satisfaction problem is deferred from the diffusion policy to another part of the robot stack, such as the low-level controller \cite{chi2023diffusion, wolf2025diffusion, trilbmteam2025carefulexaminationlargebehavior, wei2025empiricalanalysissimandrealcotraining, black2025pi05, jung2025jointmodelbasedmodelfreediffusion}. For instance, a policy operating in end-effector space may rely on a lower-level controller to enforce collision avoidance or stability constraints. Several works \cite{jung2025jointmodelbasedmodelfreediffusion,varin2019comparisonactionspaceslearning, Aljalbout_2024} have shown that the underlying control stack and action space can have a dramatic effect on policy performance. Thus, a diffusion policy's innate ability to learn and satisfy constraints in robot data is poorly understood.
In this paper, we carefully isolate and study these capabilities in the context of a bimanual manipulation task in which sampling from the constraint-satisfying set is critical for success.

\section{Central Research Question}
\label{sec:central_question}

This paper analyzes the ability of Diffusion Policy to discover a kinematic constraint present in its training dataset. We consider a bimanual pick-and-place task where the two robots coordinate to transport a box from a table onto a shelf. To complete such a task, the two end-effectors must maintain a constant relative transform. More generally, a comparable situation arises in mobile manipulation with legged robots, where stability requires maintaining a constant relative pose among the ground-contact points. In the context of our task, the kinematic constraint prevents the transported object from being dropped or pulled apart. The need to maintain this relative transform creates a nonlinear kinematic equality constraint.
Although the precise constraint changes depending on the grasp, the diffusion policy must still produce actions that lie along a manifold.

A successful demonstration requires the robots to grasp the box with both grippers and transport it to the second-highest shelf shown in \Cref{fig:teaser}.
We collect several datasets of successful demonstrations that vary in quality and quantity. We study the effect of three different factors on the policy's performance. First, we partition demonstrations into subsets and examine policy performance and constraint satisfaction as a function of dataset size. Then, we regenerate demonstrations while perturbing the robot commands during box transportation to consider the effects of constraint violations on the policy. Next, we investigate correlations between manifold curvature, policy performance, and constraint satisfaction. Finally, we translate our setup and ``transform-locking'' system to our hardware setup to examine the effectiveness of these methods in the real world.

\section{System Setup}
\label{sec:system_setup}
To collect data, we use a teleoperation stack with features that aid in collision avoidance and constraint satisfaction.

\subsection{Enforcing Kinematic Constraints}
\label{sec:system_setup:kinematic_constraints}

The user teleoperates the robot in end-effector space. To obtain the corresponding joint angles for each arm, the system employs analytic inverse kinematics, which in the case of 7-DoF manipulators includes an additional free parameter arising from the extra degree of freedom. Control over this value, which manifests visually as ``elbow-control'' over the arm, is essential for tasks that require collision avoidance~\cite{mazzaglia2024redundancy}.
We specifically use the stereographic shoulder-elbow-wrist (SSEW) angle parameterization \cite{stereographicsew}, as it avoids a ``representational'' singularity in other solutions that was problematic for our task, which our demonstrations must pass close to.
To enforce the kinematic constraint during teleoperation, the system utilizes a mode we call ``transform-locking.'' Upon entering this mode, the teleoperator chooses an arm to retain as the control arm, and the system fixes the relative pose between the end-effectors. Afterwards, the user only needs to teleoperate the control arm; the subordinate arm automatically computes and tracks IK solutions to maintain the initial relative transform between the end-effectors.
This is in the spirit of classic bimanual control algorithms~\cite{nakano1974cooperational,luh1987constrained}, and is especially useful for maintaining a constant grasp during object transportation.

Occasionally, the subordinate arm cannot perfectly track the relative transform commanded by the control arm due to singularities, joint limits, or other workspace constraints. In such cases, the control system’s transform-preservation checks detect that the command would violate the kinematic constraint, and the arm holds its most recent valid configuration until a feasible command is received.

\subsection{Teleoperation \& Data Collection}
\label{sec:system_setup:teleoperation}

The teleoperator uses a virtual reality headset with hand controllers \cite{OrbikEbert2021OculusReader} to send desired end-effector commands to the robots. To control the SSEW angle, the user moves the joystick on each controller. Additionally, the transform-locking kinematic constraint can be enabled by pressing down on the joystick of the controller corresponding to the desired control arm. This mode ensures the relative transform between end-effectors stays constant. Changes in the SSEW angle are still allowed for each arm, since this parameter does not affect the desired pose.

We used this teleoperation setup to collect a dataset for our bimanual pick-and-place task.
To start each demonstration, the box is initialized on the table in front of the robotic arms.
Its pose is drawn from the distribution $x\sim\Unif[-0.2,0.2)$, $y\sim\Unif[0.55,0.65)$, $\theta\sim\Unif[-\frac{\pi}{8},\frac{\pi}{8})$.
If the box is rotated significantly or located far from the table's center, the teleoperator demonstrates a pushing behavior to reposition the box for grasping. During the rest of the episode, the teleoperator positions the grippers around the box, locks the transform, closes the grippers, and transports the box to the shelf. An episode ends when the teleoperator releases the box, unlocks the transform, and safely navigates the arms away from the shelves. 

Both our simulation and real-world datasets were collected on two KUKA iiwa 14 robots and four cameras: two scene cameras and two wrist cameras (one per robot). The observation space of our policies includes the joint angles (7-DoF per robot), past gripper command (1-DoF per robot), and RGB images from the four cameras. The action space of our policy is a 16-dimensional vector that contains the absolute commanded joint angles and gripper actions for both robots.

We chose to output joint angle commands instead of end-effector commands for two reasons. First, the constraint manifold described in \Cref{sec:central_question} is affine in end-effector space, but highly nonlinear in configuration space. Thus, predicting actions in configuration space allows us to investigate richer and more informative questions about constraint satisfaction, such as the effect of curvature of the constraint manifold. Second, our task requires the robots to operate in and around obstacles, where pure end-effector commands often lead to collisions and poor performance~\cite{mazzaglia2024redundancy}. These problems are most naturally addressed in configuration space, where the teleoperator can leverage the kinematic redundancy of the arms to avoid collisions.

\subsection{Inference Workflow}

At inference time, the four camera feeds, robot proprioception, and gripper positions are stored as observations. The policy conditions on the last two observations and predicts the next 16 actions for the robots. The control system executes the first eight commands, interpolated with a first-order hold. Performing a first-order hold between predicted actions generates intermediate commands that might not satisfy constraints, even if the original predictions did.
Thus, in \Cref{sec:experiments}, we only evaluate constraint satisfaction at the knot-points of the policy's predicted actions.

\section{Experiments}
\label{sec:experiments}
We study how policy performance is affected by three factors: the number of demonstrations, the degree to which the demonstrations satisfy the kinematic constraints, and the curvature of the constraint manifold.
We perform extensive experiments in simulation, where these factors can be carefully controlled. Hardware experiments are used to verify the applicability of our results in the real world.

\begin{table}[t]
    \centering
    \begin{tabularx}{\columnwidth}{c c c c}
        \toprule
        \makecell{\textbf{Perturbation}\\ \textbf{Level}} & 
        \textbf{$\eta$} & 
        \makecell{\textbf{Avg. Position}\\ \textbf{Error (cm)}} & 
        \makecell{\textbf{Avg. Orientation}\\ \textbf{Error (deg)}} \\
        \midrule
        0 & 0.0    & 0.00 & 0.00 \\
        1 & 0.001  & 0.40 & 0.29 \\
        2 & 0.0025 & 1.01 & 0.71 \\
        3 & 0.005  & 2.02 & 1.43 \\
        \bottomrule
    \end{tabularx}
    \caption{This table quantifies the magnitude of the constraint violations in end-effector space for the regenerated \emph{datasets} at each perturbation level.
    The magnitude of the constraint violation in each dataset is controlled by varying the $\eta$ parameter in \Cref{eqn:sde}. The constraint violations for the policies trained on these datasets are presented in \Cref{tab:all_results}.}
    \label{tab:noise_description}
\end{table}

\subsection{Metrics for Evaluation}
\label{sec:experiments:metrics}

\begin{figure*}[h]
  \centering
    \includegraphics[width=\linewidth]{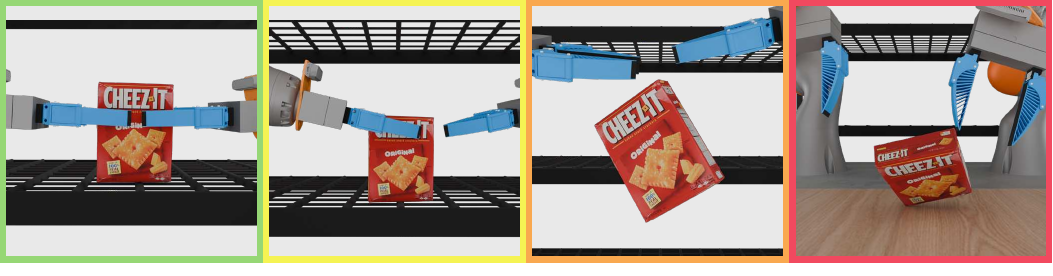}
    \vspace{-10pt}
    {
      \begingroup
      \def\W{8} %
      \def\a{25}\def\b{25}\def\c{25}\def\d{25}%
      \pgfmathsetmacro{\total}{\a+\b+\c+\d}%
      \pgfmathsetmacro{\pa}{\a/\total}%
      \pgfmathsetmacro{\pb}{\b/\total}%
      \pgfmathsetmacro{\pc}{\c/\total}%
      \pgfmathsetmacro{\pd}{\d/\total}%
      \def\H{0.4} %

      \def\aa{Full Success}
      \def\bb{Single-Gripper}
      \def\cc{Box Drop}
      \def\dd{Push Fail}
    
      Key:
      \begin{tikzpicture}[baseline=(current bounding box.center)]
        \coordinate (L) at (0,0);
        \foreach \p/\val/\col/\text in {\pa/\a/mygreen/\aa,\pb/\b/myyellow/\bb,\pc/\c/myorange/\cc,\pd/\d/myred/\dd} {
            \pgfmathsetmacro{\w}{\p*\W}
            \coordinate (R) at ($(L)+(\w,0)$);
            \fill[\col,draw=black] (L) rectangle ($(R)+(0,\H)$);
            \node[below=-2pt] at ($ (L)!0.5!(R) $) {\scriptsize \text};
            \coordinate (L) at (R);
        }
      \end{tikzpicture}%
      \endgroup
    }
    \break
    \vspace{-10pt}
  {
  \setlength\tabcolsep{2pt}
    \begin{tabularx}{\linewidth}{c c YYYY}
    & & \multicolumn{4}{c}{} \\
    & & \textbf{50 Demos} & \textbf{100 Demos} & \textbf{150 Demos} & \textbf{200 Demos} \\
    \multirow{4}{*}[-15pt]{\makebox[0pt][c]{\rotatebox{90}{\clap{\hspace{8pt}\textbf{Perturbation Level}}}}}
      & \hspace{3pt}\textbf{0} 
        & \successbarci{132}{15}{35}{18}{13.2}
        & \successbarci{137}{18}{42}{3}{12.8}
        & \successbarci{142}{27}{29}{2}{13.2}
        & \successbarci{158}{16}{23}{3}{12.4} \\
      & \hspace{3pt}\textbf{1} 
        & \successbarci{114}{21}{39}{26}{13.8}
        & \successbarci{139}{19}{34}{8}{13.8}
        & \successbarci{155}{17}{24}{2}{12.4}
        & \successbarci{151}{21}{24}{4}{12.8} \\
      & \hspace{3pt}\textbf{2} 
        & \successbarci{117}{23}{56}{4}{13.6}
        & \successbarci{125}{15}{56}{4}{13.4}
        & \successbarci{122}{23}{52}{3}{13.6}
        & \successbarci{151}{15}{32}{2}{12.8} \\
      & \hspace{3pt}\textbf{3} 
        & \successbarci{93}{32}{54}{21}{13.4}
        & \successbarci{100}{16}{77}{7}{13.4}
        & \successbarci{123}{20}{51}{6}{13.6}
        & \successbarci{118}{32}{46}{4}{13.2} \\
    \end{tabularx}
  }

    \caption{The success rates of diffusion policies trained on datasets with varying sizes and levels of perturbation. A visual depiction of each outcome is shown above. Each policy was evaluated over 200 trials; success bars include the 95\%
    Wilson Confidence Interval for the number of full successes.}
    \label{fig:sim_success_failure}
\end{figure*}

For a coarse evaluation of policy performance, rollouts of trained policies are categorized into four outcomes:
\begin{enumerate}[I.]
    \item \textbf{Full Success:} Successful grasp and transportation of the box with \textit{both} grippers;
    \item \textbf{Single-Gripper Success:} transportation of the box, but one of the grippers loses its grasp along the way;
    \item \textbf{Box Drop Failure:} Successful grasp of the box, but failure during transportation;
    \item \textbf{Full Failure}: Failure to pick up the box. 
\end{enumerate}
\Cref{fig:sim_success_failure} provides a visualization of these four categories. To produce a binary success rate, the first two outcomes are considered a ``success'' and the latter two a ``failure.''

We emphasize that these are not metrics for constraint violation, but rather metrics for task completion. For instance, a policy that cannot maintain a constant grasp may still complete the task due to compliance in the grippers, but could potentially exert damaging contact forces on the object. The latter is a consequence of constraint violation, but is not captured by our metrics for task completion.

Instead, we measure constraint violations during box transportation as the mean position and orientation error of the relative transform between the end-effectors. We choose to measure constraint violation in the end-effector space instead of configuration space since it is a reasonable proxy for constraint violation in the robot's output space, it is easier to compute, and it is more interpretable.

Specifically, for each set of grasping actions predicted by the policy, we record the relative pose between the end-effectors at the most recent observation, and then compute the position and orientation errors between this reference transform and the subsequent transforms in the predicted actions.
We measure this \textit{relative} error rather than the absolute error from the initial grasp, since the short observation history prevents the policy from accessing that information. Ultimately, the best we can expect is that the policy preserves the transform specified in its observations.
Position error is defined as the Euclidean distance between the commanded and desired locations, while orientation error is measured as the $\SO(3)$ geodesic distance between the commanded and desired rotation matrices.

Finally, we consider correlations between the policy's performance (i.e. task completion and constraint satisfaction) and the curvature of underlying constraint manifolds.
Let $\monogram{X}{L}{R}(q)$ denote the relative transform from the right to the left gripper at a configuration $q$ (using monogram notation~\cite[\S 3.1]{russtedrake2024manipulation}).
For a given trial, when the box is first grasped at configuration $q_0$, we record the transformation $\monogram{X}{L}{R}(q_0)$. The constraint manifold is then defined to be $\mc M=\set{q:\monogram{X}{L}{R}(q)=\monogram{X}{L}{R}(q_0)}$.
For a point $q\in\mc M$, we can compute the Riemannian curvature tensor $\mc R$ in terms of the Jacobian and Hessian of the function
\begin{equation}
    q\mapsto\monogram{X}{L}{R}(q)-\monogram{X}{L}{R}(q_0),
\end{equation}
which we obtain with JAX~\cite{jax2018}.
We then take the squared Frobenius norm of $\mc R$, also called the Kretschmann scalar~\cite{henry2000kretschmann}, to quantify how much the manifold differs from Euclidean space with a single scalar\footnote[1]{
    The more commonly used ``scalar curvature'' averages positive and negative values, so it can experience cancellation.
}.

When $q$ is only near $\mc M$ (due to slight constraint violations in the policy), we instead compute the curvature $\tilde{\mc R}$ of $\tilde{\mc M}=\set{\tilde q:\monogram{X}{L}{R}(\tilde q)=\monogram{X}{L}{R}(q)}$.
A simple perturbation analysis shows that the error in the Kretschmann scalar varies quadratically with the distance from the manifold, so our computations are still reasonable.

\subsection{Simulation Experiments}
\label{sec:experiments:simulation}

In total, the teleoperator collected 200 demonstrations of the pick-and-place task using the ``transform-locking" mechanism to satisfy the kinematic constraint. With transform-locking enabled, the Euclidean error from the desired position was less than $10^{-10}$ meters; similarly, the $\SO(3)$ rotation distance from the desired orientation was less than $10^{-6}$ radians. These tolerance levels were treated as ``no constraint violation." To train more robust policies, we first collected 100 demonstrations and identified common failure modes of the resulting policy. We then patched these modes in the final dataset by providing short recovery episodes.

By varying the number of demonstrations, we examine how dataset size influences the policy’s success rate and its tendency to violate constraints. We further analyze how these outcomes depend on dataset quality, while noting that additional factors, such as gripper compliance, also affect the overall success rate. More concretely, we analyze how the magnitude of constraint violations in the demonstrations influences downstream policy performance. To simulate demonstrations with greater violations, we inject correlated random noise into the end-effector commands at each timestep $t$ according to the random process $Z_t$. We define $Z_t$ by the stochastic differential equation in Eq. \eqref{eqn:sde}, where $W_t$ is a Wiener process. We sample realizations of $Z_t$ using the Euler-Maruyama method \cite{kloeden_platen_1992}.
\begin{equation}
\begin{aligned}
\mathrm{d}Z_t &= -\alpha Z_t\mathrm{d}t + \eta \, \mathrm{d}W_t, \\
   &\;\;\eta > 0, \  \alpha = 0.01,\  Z_0 = 0
\end{aligned}
\label{eqn:sde}
\end{equation}
Perturbing the clean data with Eq. \eqref{eqn:sde} (as opposed to i.i.d. noise) allows us to correlate the constraint violations across time. This provides a more realistic model for low-quality data. We vary the magnitude of the constraint violation by setting the volatility, $\eta$, to four different values. The resulting mean position and orientation errors for the datasets at four ``perturbation levels'' are computed in \Cref{tab:noise_description}. We chose these levels through progressive scaling of the $\eta$ parameter until regenerated demonstrations consistently failed the task. In other words, the highest perturbation level represents the lowest-quality data that can still feasibly accomplish the task.

 This process allows us to measure the performance of the policy as a function of the quantity and quality of demonstrations. We train policies for varying numbers of demonstrations and amounts of constraint violation.  During policy evaluation, we run 200 trials with the box's initial pose $(x,y,\theta)$ drawn according to $x\sim\Unif[-0.1,0.1)$, $y\sim\Unif[0.575,0.625)$, and $\theta\sim[-\frac{\pi}{16},\frac{\pi}{16})$. We record the success category of each trial and report results in \Cref{fig:sim_success_failure}. Besides task success rates, the performance of each policy is evaluated according to two additional metrics: relative transform position error and relative transform orientation error (see \Cref{tab:all_results}).
For all 16 policies, we consider correlations between constraint error calculated at a timestep and the curvature of the constraint manifold at the corresponding robot configuration.
In addition, we evaluate whether rollout-level curvature statistics (mean and maximum) carry predictive information about task outcome categories by computing the Jensen–Shannon divergence \cite{Lin1991JensenShannon} between the corresponding conditional distributions.

\subsection{Hardware Experiments}
\label{sec:experients:hardware}

\begin{figure}[t]
    \centering
    \includegraphics[width=\linewidth]{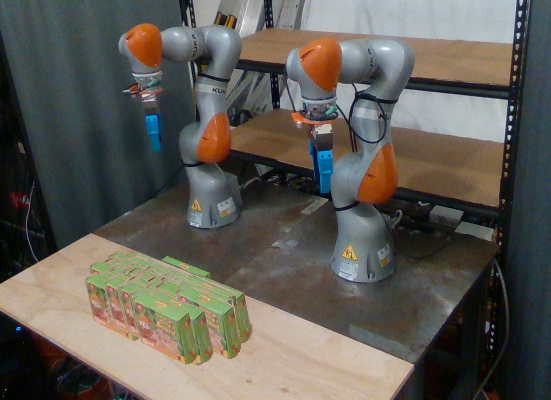}
    \caption{An overlay of the initial conditions for the policy evaluations performed on the hardware bimanual setup.}
    
    \label{fig:hardware_overlay}
\end{figure}

For the hardware dataset, 100 demonstrations of the box task were collected. After performing initial rollouts, we examined common failure cases and added or removed demonstrations as needed from the set of 100. We trained a policy on the finalized dataset and performed evaluations according to a box initialization distribution similar to that of the training distribution. A visual depiction of the evaluation distribution is shown in \Cref{fig:hardware_overlay}. The success rates, including categorization, and the position and orientation error from the kinematic constraint transform can be seen in \Cref{tab:hardware_success}.

\section{Results}
\label{sec:results}
\newcommand{\stackbar}[4]{%
    #1\textbackslash#2\textbackslash#3\textbackslash#4%
}

\begin{table*}[h]
    \centering

  \setlength\tabcolsep{2pt}
  
\begin{tabularx}{0.495\textwidth}{cYYYY}
    \toprule
    \multicolumn{5}{c}{\textbf{Average Position Error (cm)}} \\
    \midrule
    \textbf{Perturbation} & \textbf{50 Demos} & \textbf{100 Demos} & \textbf{150 Demos} & \textbf{200 Demos} \\
    \midrule
    Level 0 & $1.38 \pm 0.83$ & $1.32 \pm 0.71$ & $1.24 \pm 0.61$ & $1.22 \pm 0.58$ \\
    Level 1 & $1.44 \pm 0.86$ & $1.35 \pm 0.72$ & $1.28 \pm 0.62$ & $1.22 \pm 0.57$ \\
    Level 2 & $1.61 \pm 0.92$ & $1.36 \pm 0.70$ & $1.27 \pm 0.61$ & $1.27 \pm 0.59$ \\
    Level 3 & $1.74 \pm 1.02$ & $1.48 \pm 0.74$ & $1.44 \pm 0.68$ & $1.30 \pm 0.60$ \\
    \bottomrule
\end{tabularx}
\hfill
\begin{tabularx}{0.495\textwidth}{cYYYY}
    \toprule
    \multicolumn{5}{c}{\textbf{Average Orientation Error (degrees)}} \\
    \midrule
    \textbf{Perturbation} & \textbf{50 Demos} & \textbf{100 Demos} & \textbf{150 Demos} & \textbf{200 Demos} \\
    \midrule
    Level 0 & $2.04 \pm 1.47$ & $2.00 \pm 1.12$ & $1.88 \pm 0.91$ & $1.80 \pm 0.85$ \\
    Level 1 & $2.13 \pm 1.45$ & $2.00 \pm 1.14$ & $1.88 \pm 0.92$ & $1.83 \pm 0.90$ \\
    Level 2 & $2.48 \pm 1.48$ & $1.91 \pm 1.01$ & $1.91 \pm 0.94$ & $1.82 \pm 0.88$ \\
    Level 3 & $2.36 \pm 1.51$ & $2.05 \pm 1.07$ & $2.06 \pm 1.00$ & $1.87 \pm 0.87$ \\
    \bottomrule
\end{tabularx}

    \vspace{\baselineskip}
    \includegraphics[width=\linewidth]{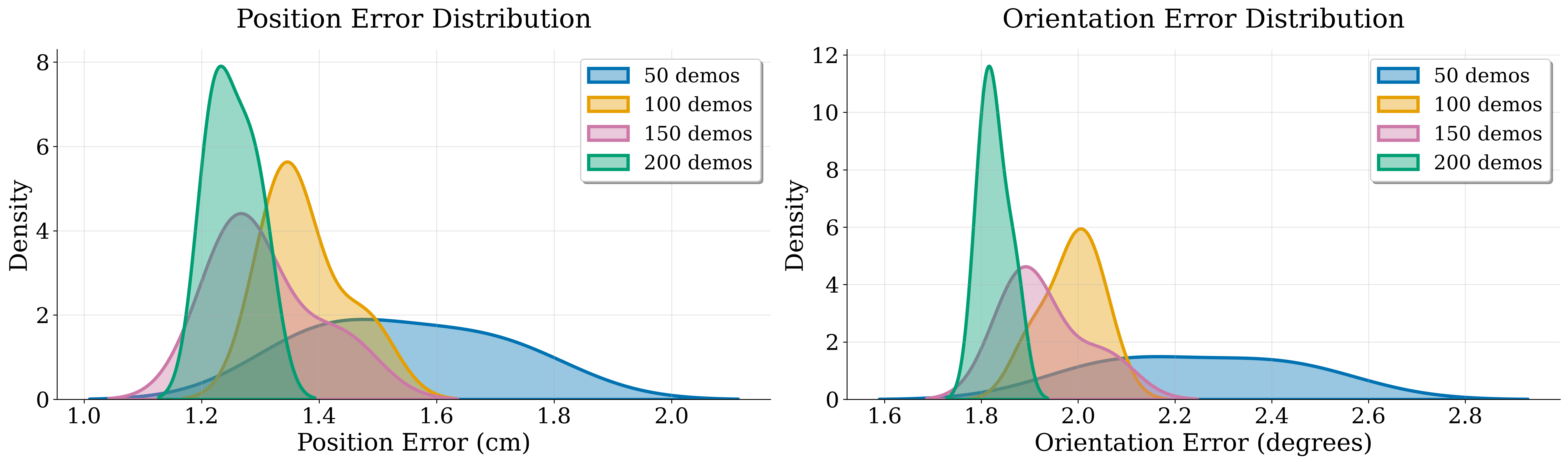}

    \caption{Performance metrics of diffusion policies trained on datasets of different sizes with varying amounts of perturbation. The top tables show the average ($\pm$ standard deviation) position and orientation error in the end-effectors' relative transform for each policy's predicted actions. These tables quantify the magnitude of the constraint violation in end-effector space. The bottom plots visualize the distribution of the errors at each data scale.\vspace{-\baselineskip}}
    \label{tab:all_results}
\end{table*}

We use the results from \Cref{sec:experiments} to answer the following questions about policy learning on constraint manifolds.
\begin{enumerate}
    \item How do the quantity and quality of the data affect the policy's task completion rates?
    \item How do the quantity and quality of the data affect the policy's constraint satisfaction?
    \item How does the curvature of the manifold affect the policy's performance?
\end{enumerate}
We also analyze the performance of our policy on hardware.

\subsection{Simulation Results}

\subsubsection{Success Rates}
The results from our simulation experiments are presented in \Cref{tab:all_results}. Overall, the task completion rates of the policies followed the expected ``scaling laws’’ with respect to the dataset size and quality. For example, the task completion rates at all perturbation levels improved when policies were trained with more data.

While both data quantity and quality affected the policy’s performance, our results suggest that in the single-task setting, the quality of the training data is a stronger predictor of task completion than data quantity. Notably, the policy trained with just 50 clean demonstrations achieved a higher “Full Success” rate than the policy trained with 200 demonstrations at the highest perturbation level. \Cref{tab:all_results} shows this trend more generally: policies trained with greater levels of constraint violation were less likely to complete the task. This suggests that for tasks where success depends on satisfying a constraint, high-quality demonstrations that lie along the constraint manifold are important for effective policy learning.

Interestingly, the success rates for policies trained at perturbation level 1 performed on par with or slightly better than the policies trained on data with perfect constraint satisfaction. This result could arise from a few factors. One potential explanation is that mild data noising can expand the support of the dataset, resulting in policies that are more robust to distribution shift during deployment. In other words, the benefits from greater diversity in the training data at low perturbation levels may have outweighed the effect of minor constraint violations. These observations are consistent with theoretical results from previous work \cite{block2023provableguaranteesgenerativebehavior}.

\subsubsection{Constraint Satisfaction}
Constraint satisfaction improves with more data and cleaner data, but appears to be more sensitive to dataset size than to noise level. We highlight the following two observations.

\textbf{Diffusion policies struggle to sample \emph{exactly} from the kinematic constraint manifold}: even the policy trained with 200 clean demonstrations exhibits 1.22cm and $1.80$\textdegree{} of average error in the relative transform constraint. Nevertheless, diffusion policies can \textbf{learn the coarse structure of the manifold given sufficient data}, even when this data only \textit{approximately} satisfies the constraints. For example, when the clean actions were noised to the highest perturbation levels, the position error of the policies trained with 200 demos only increased by 6.6\%. On the other hand, the error of the policies trained with just 50 demos increased by 26\%.

This suggests that with enough data, real-world diffusion models could be ``averaging out'' constraint-violating samples to recover the data manifold’s approximate structure. This echoes theoretical findings in classical statistics. \cite{belkin2018doesdatainterpolationcontradict} showed that even interpolating estimators--which exactly fit (potentially noisy) samples--can still achieve minimax-optimal generalization rates. While the translation between our setting and the theorems in \cite{belkin2018doesdatainterpolationcontradict} is not exact, the qualitative lesson carries over: with sufficient data and model capacity, fitting labels with zero-mean additive noise does not preclude learning the underlying manifold.

These observations also imply that in the low-data regime (where robotics typically operates), data quality is especially impactful. In contrast, when data is plentiful, even imperfect demonstrations can help the model infer the constraint.

\subsubsection{Curvature}

Across the board, we found almost no correlation between the Kretschmann scalar and error.
The average Pearson correlation coefficient across all 16 policies was $-0.098 \pm 0.033$, with values ranging from $-0.165$ to $-0.036$. Similarly, the average Spearman correlation coefficient was $-0.110 \pm 0.036$, with values from $-0.185$ to $-0.046$. For both metrics, these values indicate a negligible amount of correlation.

From a broader perspective, we investigated whether the success of a given policy rollout can be easily predicted based on the curvature it experienced. In this case as well, we found that both the mean and maximum curvature during a rollout provided virtually no predictive power.

Formally, let $K_\mathrm{mean}$ and $K_\mathrm{max}$ be random variables that describe the mean and maximum Kretschmann scalar during a rollout; let $R$ be the result of the rollout. We use the policy rollouts to construct a kernel density estimate for the distributions $K_\mathrm{mean} | R={\tt Full\ Success}$, $K_\mathrm{mean} | R={\tt Single\ Gripper}$, and $K_\mathrm{mean} | R={\tt Box\ Drop}$\footnote[2]{
    We do not include $K_\mathrm{mean} | R={\tt Failure}$ since the constrained motion never occurs in this case, and so the curvature analysis is not applicable.
} with bandwidth chosen according to Scott's Rule~\cite{scott2015multivariate} (and likewise for $K_\mathrm{max}$).
We are interested in computing the three-way Jensen Shannon (JS) divergence for the three distributions for mean curvature \cite{Lin1991JensenShannon}, and likewise for the maximum curvature.

On a per-policy basis, we found very little divergence between the curvature distributions. For mean and maximum curvature, the highest divergence observed was 0.08 and 0.06 nats, respectively. Similarly, the average JS divergence across policies for mean and maximum curvature was 0.04 and 0.03 nats. The low JS divergence values imply that the curvature of a rollout was not an informative statistic for success rate. For reference, the maximum possible divergence, indicating completely disjoint distributions, is $\ln(3)=1.099$ nats.

Clearly, the policy success rate, and even the low-level incremental error, is independent of the curvature experienced along a given rollout. We propose two possible explanations.
One is that curvature still plays a role, but it is not revealed in the Krestchmann scalar.
We know that the rollouts only explore a small portion of the whole configuration space, so perhaps the curvature along some smaller submanifold that contains the rollouts is more descriptive.
A second possible explanation is that the geometry of the \emph{data} manifold is playing a much larger role in the performance of the diffusion policy.
But since we only know a coarse approximation of the data manifold from our limited demonstrations, measuring its curvature is intractable.

\subsection{Hardware Experiments}

As seen in \Cref{tab:hardware_success}, the hardware policy succeeded at the task $73\%$ of the time. Box pushing presented a much larger challenge on hardware than in simulation, with this category making up nearly a fifth of all evaluation rollouts. The box used in the hardware dataset was slim and light; thus, it was easy for the robotic arms to accidentally topple it during this portion of the task. Box dropping contributed significantly less to the failure rate than in simulation; in fact, most box drops were a result of missing the shelf due to a poor initial grasp as a result of planar pushing error. Similarly, ``Single-Gripper Success" made up $13\%$ of all evaluations and were often results of off-center grasps caused by faulty planar pushing. 

Both the simulated and real-world policies exhibited constraint violation on the order of 1-3cm and 2-2.5\textdegree{}. Nonetheless, many of the policies still completed the task reliably. We observed that compliant grippers and strong grasping forces helped mitigate the effects of minor constraint violations during object transport. This illustrates that effective low-level control and compliance along with algorithmic advances in imitation learning are both important for high success rates in robotics.

\begin{table}[t]
    \centering
    \begin{tabular}{ccccccc}
        \toprule
        \makecell{\textbf{Overall} \\ \textbf{Task} \\ \textbf{Success}} & 
        \textbf{I} & 
        \textbf{II} & 
        \textbf{III} & 
        \textbf{IV} & 
        \makecell{\textbf{Avg. Pos.} \\ \textbf{Error} \\ \textbf{(cm)}} & 
        \makecell{\textbf{Avg. Ori.} \\ \textbf{Error} \\ \textbf{(deg)}} \\
        \midrule
        $73\%$ & $60\%$ & $13\%$ & $8\%$ & $19\%$ & $0.81 \pm 0.42$ & $1.04 \pm 0.55$ \\
        \bottomrule
    \end{tabular}
    \caption{The task completion rates of a diffusion policy trained on 100 hardware demonstrations for the bimanual pick-and-place task. Success rates are categorized into four different outcomes, I--IV, as described in \Cref{sec:experiments}. Overall, the policy achieves a binary success rate of 73\%. The last two columns show the position and orientation error for the kinematic constraint.}
    \label{tab:hardware_success}
\end{table}

\section{Conclusion}
\label{sec:conclusion}
We draw two main conclusions, and make one conjecture that requires further study to confirm:
\begin{enumerate}
    \item Diffusion policies learn only a coarse approximation of kinematic constraint manifolds, even when trained on perfectly constraint-satisfying data, yet they can still complete tasks successfully. This robustness is likely attributable to compliance, either from robust low-level controllers or from physically compliant grippers.
    \item Data quality is a critical determinant of diffusion policy success when kinematic constraints are involved. In our experiments, datasets with an average of just 1.4cm positional error and 2\textdegree{} orientation error required four times as many demonstrations to achieve the same success rate as perfectly constraint-satisfying data. These results underscore the importance of collecting high-quality teleoperation data.
    \item \emph{We conjecture that} even when the training data are perturbed off the constraint manifold by an approximately zero-mean process (such as teleoperator error), a diffusion policy can still recover the coarse structure of the manifold given sufficient data.
\end{enumerate}

Future work should investigate whether these conclusions hold across a broader range of kinematically constrained tasks and under different action spaces, such as end-effector poses or relative actions. It also remains an open question whether these tendencies are unique to diffusion policies or apply more broadly to imitation learning. Finally, we hope that core elements of our experimental framework, such as the ``constraint-locking'' teleoperation mode and controlled data perturbations, will inspire further studies on constraint satisfaction in robot imitation learning.

\bibliographystyle{IEEEtran}
\bibliography{ref.bib}

\end{document}